\documentclass[10pt,twocolumn,letterpaper]{article}

\usepackage{cvpr}
\usepackage{times}
\usepackage{epsfig}
\usepackage{graphicx}
\usepackage{amsmath}
\usepackage{amssymb}
\usepackage{mathtools}
\usepackage{multirow}
\usepackage{colortbl}
\usepackage{xfrac}

\newcommand{\norm}[1]{\left\lVert#1\right\rVert}
\newcommand{\parens}[1]{\left(#1\right)}

\definecolor{Yellow}{rgb}{1,1, 0.7}

\usepackage[breaklinks=true,bookmarks=false]{hyperref}

\cvprfinalcopy % *** Uncomment this line for the final submission

\def\cvprPaperID{3751} % *** Enter the CVPR Paper ID here

\ifcvprfinal\fi
\begin{document}

%%%%%%%%% TITLE
\title{Aperture Supervision for Monocular Depth Estimation}
\author{Pratul P. Srinivasan$^1$ \thanks{Work done while interning at Google Research.}
\and
Rahul Garg$^2$
\and 
Neal Wadhwa$^2$
\and
Ren Ng$^1$
\and
Jonathan T. Barron$^2$\\
\and
$^1$UC Berkeley, $^2$Google Research
}

\maketitle
\thispagestyle{empty}

%%%%%%%%% ABSTRACT
\begin{abstract}
We present a novel method to train machine learning algorithms to estimate scene depths from a single image, by using the information provided by a camera's aperture as supervision. Prior works use a depth sensor's outputs or images of the same scene from alternate viewpoints as supervision, while our method instead uses images from the same viewpoint taken with a varying camera aperture. To enable learning algorithms to use aperture effects as supervision, we introduce two differentiable aperture rendering functions that use the input image and predicted depths to simulate the depth-of-field effects caused by real camera apertures. We train a monocular depth estimation network end-to-end to predict the scene depths that best explain these finite aperture images as defocus-blurred renderings of the input all-in-focus image.
\end{abstract}

%%%%%%%%% BODY TEXT
\section{Introduction}

The task of inferring a 3D scene from a single image is a central problem in human and computer vision. In addition to being of academic interest, monocular depth estimation also enables many applications in fields such as robotics and computational photography. Currently, there are two dominant strategies for training machine learning algorithms to perform monocular depth estimation: direct supervision and multi-view supervision.
Both approaches require large datasets where varied scenes are imaged or synthetically rendered.
In the direct supervision strategy, each scene in the dataset consists of a paired RGB image and ground truth depth map (from a depth sensor or a rendering engine), and an algorithm is trained to regress from each input image to its associated ground truth depth.
In the multi-view supervision strategy, each scene in the dataset consists of a pair (or set) of RGB images of the same scene from different viewpoints, and an algorithm is trained to predict the depths for one view of a scene that best explain the other view(s) subject to some geometric transformation. 
Both strategies present significant challenges.
The depth sensors required for direct supervision are expensive, power-hungry, low-resolution, have limited range, often produce noisy or incomplete depth maps, usually work poorly outdoors, and are  challenging to calibrate and align with the ``reference'' RGB camera. Multi-view supervision ameliorates some of these issues but requires at least two cameras or camera motion, and has the same difficulties as classic stereo algorithms on image regions without texture or with repetitive textures.

\begin{figure}
\begin{center}
\newcommand{\width}{1.0\linewidth}
\includegraphics[width=\width]{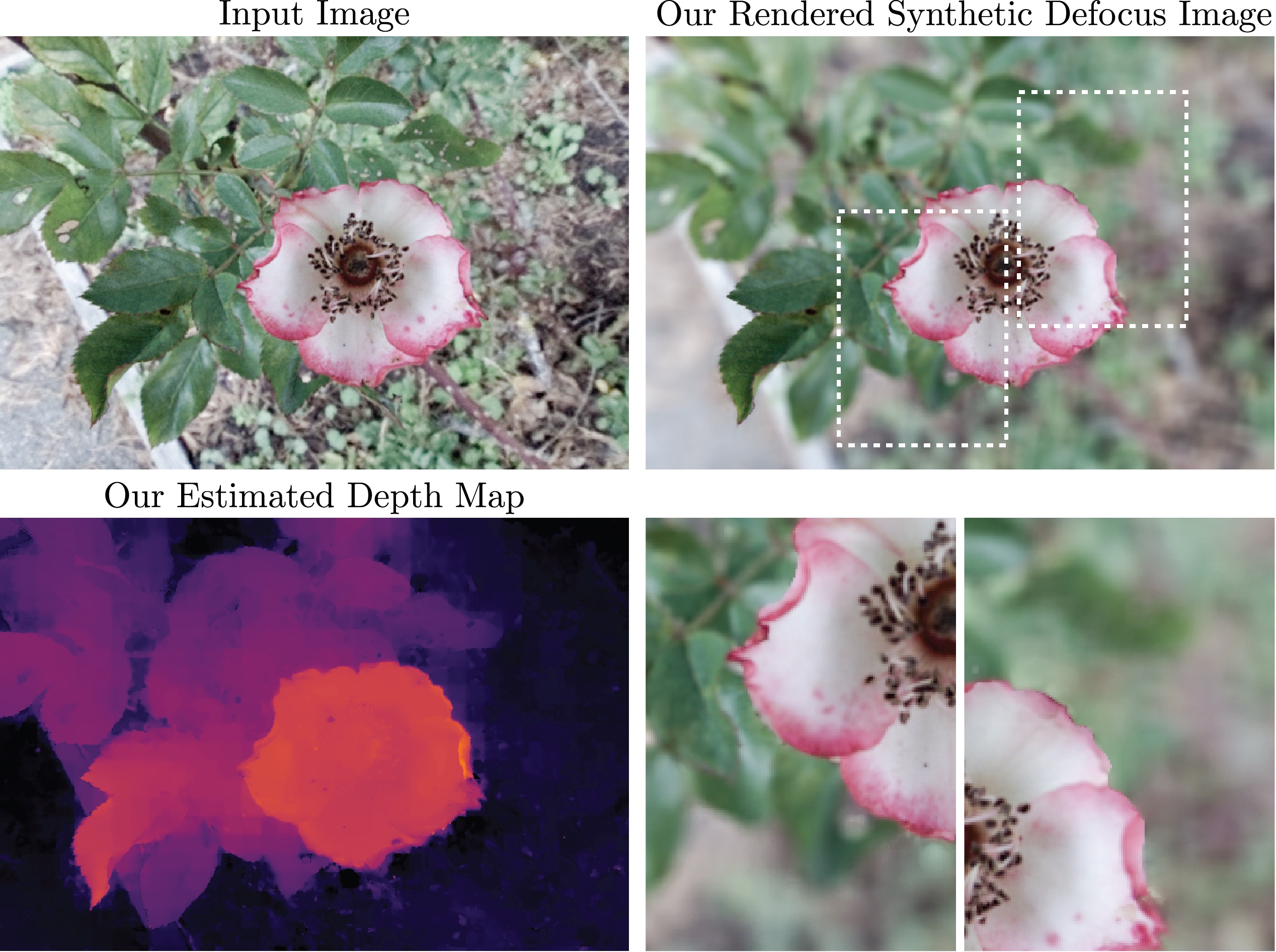} 
\caption{Given a single all-in-focus image, our algorithm estimates a depth map of the scene using a monocular depth estimation network. The only supervisory signal used to train this network was images taken from a single camera with different aperture sizes. This ``aperture supervision'' allows for diverse monocular depth estimation datasets to be gathered more easily. Depth-estimation models trained using aperture supervision estimate depths that work particularly well for generating images with synthetic shallow depth-of-field effects.}
\label{fig:teaser}
\end{center}
\vspace{-0.2in}
\end{figure}

In this work, we propose a novel strategy for training machine learning algorithms to perform monocular depth estimation: \emph{aperture supervision}.
We demonstrate that sets of images taken by the same camera and from the same viewpoint but with different aperture sizes can be used to train a monocular depth estimation algorithm.
Aperture supervision can be used for general-purpose monocular depth estimation, but works particularly well for one compelling computational photography application: synthetic defocus.
This is because the algorithm is trained end-to-end to predict scene depths that best render images with defocus blur; the loss used during training is exactly consistent with the task in question.
Figure~\ref{fig:teaser} shows an example input all-in-focus image, and our algorithm's predicted depth map and rendered shallow depth-of-field image.

% \paragraph{Aperture Supervision}
An image taken with a small camera aperture (e.g. a pinhole) has a large depth-of-field, causing all objects in the scene to appear sharp and in focus.
If the same image is instead taken with a larger camera aperture, the image has a shallow depth-of-field, and objects at the focal plane appear sharp while other objects appear more blurred the further away they are from the focal plane.
We exploit this depth-dependent difference between images taken with smaller and larger apertures to train a convolutional neural network (CNN) to predict the depths that minimize the difference between the ground truth shallow depth-of-field images and shallow depth-of-field images rendered from the input all-in-focus image using the predicted depths.

% \paragraph{Differentiable Aperture Rendering}
To train an end-to-end machine learning pipeline using aperture supervision, we need a differentiable function to render a shallow depth-of-field image from an all-in-focus image and a predicted depth map.
In this work we propose two differentiable aperture rendering functions (Section~\ref{sec:rendering}).
Our first approach, which we will call the ``light field'' model, is based on prior insights regarding how shearing a light field induces focus effects in images integrated from that light field.
Our light field model uses a CNN to predict a depth map that is then used to warp the input 2D all-in-focus image into an estimate of the 4D light field inside the camera, which is then focused and integrated to render a shallow depth-of-field image of the scene.
Our second approach, which we will call the ``compositional'' model, eschews the formal geometry of image formation with regards to light fields, and instead approximates the shallow depth-of-field image as a depth-dependent composition of blurred versions of the all-in-focus image.
Our compositional model uses a CNN to predict a probabilistic depth map (a probability distribution over a fixed set of depths for each pixel) and renders a shallow depth-of-field image as a composition of the input all-in-focus image blurred with a representative kernel for each discrete depth, blended using the probabilistic depth map as weights.
Both of these approaches allow us to express arbitrary aperture sizes, shapes, and distances from the camera to the focal plane, but each approach comes with different strengths and weaknesses, as we will show.

\section{Related Work}
\paragraph{Inferring Geometry from a Single Image}
Early works in computer vision such as  shape-from-shading~\cite{Horn75,Zhang99} and shape-from-texture~\cite{Malik94,Witkin81} exploit specific cues and explicit knowledge of imaging conditions to estimate object geometry from a single image.
The work of Barron and Malik~\cite{Barron15} tackles a general inverse rendering problem and recovers object shape, reflectance, and illumination from a single image by solving an optimization problem with priors on each of these unknowns.
Other works pose monocular 3D recovery as a supervised machine learning problem, and train models to regress from an image to ground truth geometry obtained from 3D scanners, depth sensors, or human annotations~\cite{Eigen15,Hoiem2005,Saxena08}, or datasets of synthetic 3D models~\cite{Choy16,Fan17}.

These ground truth datasets are typically low-resolution and are difficult to gather, especially for natural scenes, so recent works have focused on training geometry estimation algorithms without any ground-truth geometry.
One popular strategy for this is multi-view supervision: the geometry estimation networks are trained by minimizing the expected loss of using the predicted geometry to render ground truth views from alternate viewpoints.
Many successful monocular depth estimation algorithms have been trained in this fashion using calibrated stereo pairs~\cite{Garg16,Godard17,Xie16}.
The work of Tulsiani \etal~\cite{Tulsiani17} proposed a differentiable formulation of consistency between 2D projections of 3D voxel geometry to predict a 3D voxel representation from a single image using calibrated multi-view images as supervision.
Zhou \etal~\cite{Zhou17} relaxed the requirement of calibrated input viewpoints to train a monocular depth estimation network with unstructured video sequences by estimating both scene depths and camera pose.
Srinivasan \etal~\cite{Srinivasan17} used plenoptic camera light fields as dense multi-view supervision for monocular depth estimation, and demonstrated that the reconstructed light fields can be used for applications such as synthetic defocus and image refocusing.
In contrast to these methods, our monocular depth estimation algorithm can be trained with sets of images taken from a single viewpoint with different aperture settings on a conventional camera, and does not require a moving camera, a stereo rig, or a plenoptic camera.
Furthermore, our algorithm is trained end-to-end to estimate depths that are particularly suited for the application of synthetic defocus, much like how multi-view supervision approaches are well-suited to view-synthesis tasks.

\begin{figure*}
\begin{center}
\newcommand{\width}{1.0\linewidth}
\includegraphics[width=\width]{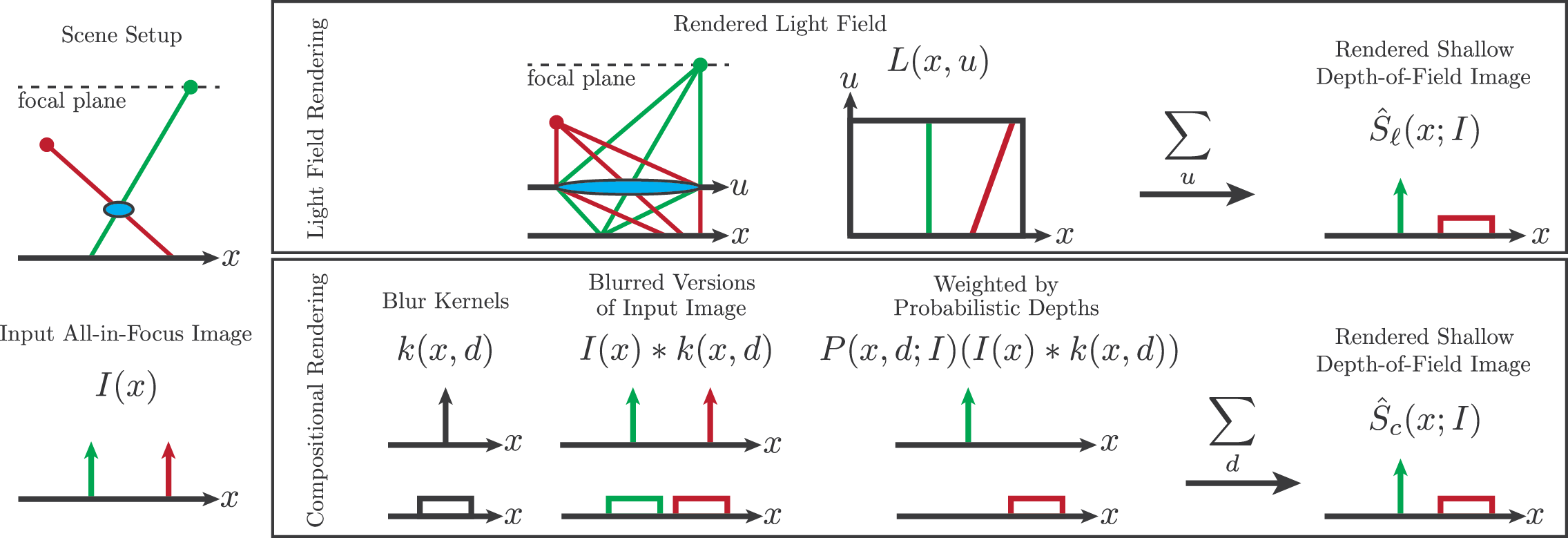} 
\caption{An illustration of our ``light field'' and ``compositional'' aperture rendering functions on a toy 1-D scene, consisting of 2 diffuse points (red and green circles) at different depths. In the input all-in-focus image, imaged through a small aperture (blue ellipse), both scene points are imaged to delta functions on the image plane (black line). The ``light field'' rendering function (top) takes this image and a depth map of the scene as inputs, predicts the light field within a virtual camera with a finite sized aperture, and integrates the rays across this entire aperture to render a shallow depth-of-field image. The ``compositional'' rendering function (bottom) takes the all-in-focus image and a probability mass function over a discrete set of depths for each pixel, and renders the shallow depth-of-field image by blending the input image blurred with a disk kernel corresponding to each discrete depth, weighted by the probability of each depth.
}
\label{fig:toy_rendering}
\end{center}
\vspace{-0.2in}
\end{figure*}

\vspace{-0.1in}
\paragraph{Light Fields} The 4D light field~\cite{Levoy96} is the total spatio-angular distribution of light rays passing through a region of free space.
Previous work has shown that pinhole images from different viewpoints are equivalent to 2D slices of the 4D light field~\cite{Levoy96}, and that a photograph with some desired focus distance and aperture size can be rendered by integrating a sheared 4D light field \cite{Isaksen00,Levoy96,Ng05}.
Our work makes use of these fundamental observations about light fields and embeds them into a machine learning pipeline to differentiably render shallow depth-of-field images, thus enabling the use of aperture effects as a supervisory signal for training a monocular depth estimation model.

\paragraph{Synthetic Defocus} Rendering depth-of-field effects is important for generating realistic imagery, and synthetic defocus has been of great interest to the computer graphics community~\cite{Cook84,Hammon,Yu10}.
These techniques assume the scene geometry, reflectance properties, and lighting are known, so other works have addressed the rendering of depth-of-field effects from the relatively limited information present in captured images.
These include techniques such as magnifying the amount of defocus blur already present in a photograph~\cite{Bae07}, using stereo to predict disparities for rendering synthetic defocus~\cite{Barron2015A}, using multiple input images taken with varying focus distances~\cite{Jacobs12} or aperture sizes~\cite{Hasinoff07}, and relying on semantic segmentation to estimate and defocus the background of monocular images~\cite{GooglePortraitMode}.
In contrast to these methods, we focus on using depth-of-field effects as a supervisory signal to train machine learning algorithms to estimate depth from a single image, and our method does not require multiple input images, external semantic supervision, or any measurable defocus blur in the input image.

\section{Differentiable Aperture Rendering}
\label{sec:rendering}

To utilize the depth-dependent differences between an all-in-focus image and large-aperture image as a supervisory signal to train a machine learning model, we need a differentiable function for rendering a shallow depth-of-field image from an all-in-focus image and scene depths (we use ``depth'' and ``disparity'' interchangeably to refer to disparity across a camera's aperture).

The depth-of-field effect is due to the fact that the light rays emanating from points in a scene are distributed over the entirety of a camera's aperture. 
Rays that originate from points on the focal plane are focused into points on the image sensor, while rays from points at other distances converge in front of or behind the sensor, resulting in a blur on the image plane. 
In this section, we present two models of this effect: a ``light field'' aperture rendering function that models the light field within a camera, and a ``compositional'' model that treats defocus blur as a blended composition of the input image convolved with differently-sized blur kernels. 
These operations both take as input an all-in-focus image and some representation of scene depth, and produce as output a rendered shallow depth-of-field image (Figure~\ref{fig:toy_rendering}).
In Section~\ref{sec:depth_estimation}, we will describe how these functions can be integrated into learning pipelines to enable aperture supervision --- the end-to-end training of a monocular depth estimation network using only shallow depth-of-field images as a supervisory signal.

\subsection{Light Field Aperture Rendering}
\label{sec:light_field}

% \paragraph{Predicting the Camera Light Field}

Our light field aperture rendering function takes as input an all-in-focus image and a depth map of the scene, and renders the corresponding shallow depth-of-field image.
This rendering function is differentiable with respect to the all-in-focus image and depth map used as input.
The rendering works by using the depth map to warp the input image into all the viewpoints in the camera light field that we wish to render.
Forward warping, or ``splatting'', the input image into the desired viewpoints based on the input depth map would produce holes in the resulting light field and consequently produce artifacts in the output rendering.
Therefore, we use a CNN $g(\cdot)$ with parameters $\boldsymbol\theta_e$ that takes the single input depth map $Z(\mathbf{x};I)$ and expands it into a depth map $D(\mathbf{x}, \mathbf{u})$ for each view in the light field:
\begin{equation}
\label{eq:depth_expansion}
D(\mathbf{x},\mathbf{u})=g_{\boldsymbol\theta_e}(Z(\mathbf{x};I))
\end{equation}
where $\mathbf{x}$ are spatial coordinates of the light field on the image plane and $\mathbf{u}$ are angular coordinates of the light field on the aperture plane (equivalent to the coordinates of the center of projection of each view in the light field).
Note that we consider the input depth map and all-in-focus image $I(\mathbf{x})$ as corresponding to the central view ($\mathbf{u}=\mathbf{0}$) of the light field.
%Because splatting into the $\mathbf{u} = \mathbf{0}$ view of the light field is a no-op and should therefore not create any holes, the $g(\cdot)$ that is learned causes $D(\mathbf{x},\mathbf{0}) \approx Z(\mathbf{x};I)$.

We use these depth maps to warp the input all-in-focus image to every view of the light field in the camera by:
\begin{equation}
\label{eq:warp}
L(\mathbf{x},\mathbf{u})=I(\mathbf{x}+\mathbf{u}D(\mathbf{x},\mathbf{u}))
\end{equation} 
where $L(\mathbf{x},\mathbf{u})$ is the simulated camera light field.

% \paragraph{Rendering a Shallow Depth-of-Field Image}

\begin{figure*}
\begin{center}
\newcommand{\width}{1.0\linewidth}
\includegraphics[width=\width]{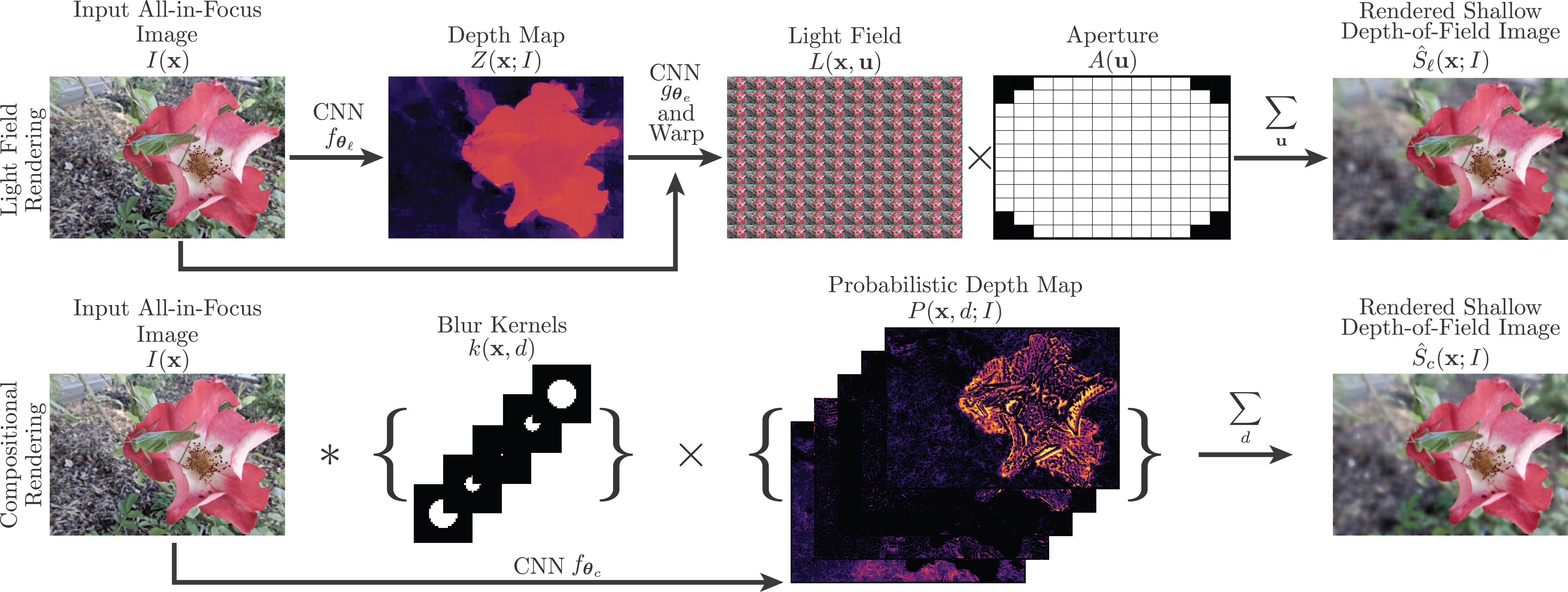} 
\caption{An overview of the full monocular depth estimation pipeline for both aperture rendering functions. When using the light field model, CNN $f_{\boldsymbol\theta_\ell}(\cdot)$ is trained to predict a depth map from the input all-in-focus image, CNN $g_{\boldsymbol\theta_e}(\cdot)$ expands this depth map into a depth map for each view, the camera light field is rendered by warping the input image into each view using the expanded depth maps, and finally all views in the light field are integrated to render a shallow depth-of-field image. When using the compositional model, the input all-in-focus image is convolved with a discrete set of disk blur kernels, and CNN $f_{\boldsymbol\theta_c}(\cdot)$ predicts a probabilistic depth map that is used to blend these blurred images into a rendered shallow depth-of-field image.}
\label{fig:pipelines}
\end{center}
\vspace{-0.2in}
\end{figure*}

After rendering the camera light field, we shear the light field to focus at the desired depth in the scene, and add the rays that arrive at each sensor pixel from across the entire aperture to render a shallow depth-of-field image $\hat{S}_\ell(\mathbf{x}; I, \hat{d})$ focused at a particular depth $\hat{d}$:
\begin{equation}
\label{eq:shear_project}
\hat{S}_\ell(\mathbf{x}; I, \hat{d}) = \sum_{\mathbf{u}}A(\mathbf{u}) L(\mathbf{x}+\mathbf{u} \hat{d}, \mathbf{u})
\end{equation} 
where $A(\mathbf{u})$ is an indicator function for the disk-shaped camera aperture that takes the value $1$ for views within the camera's aperture and $0$ otherwise. Figure~\ref{fig:pipelines} illustrates how the rendered light field is multiplied by $A(\mathbf{u})$ and integrated to render a shallow depth-of-field image.

\subsection{Compositional Aperture Rendering}
\label{sec:composition}

\begin{figure}
\begin{center}
\newcommand{\width}{1.0\linewidth}
\includegraphics[width=\width]{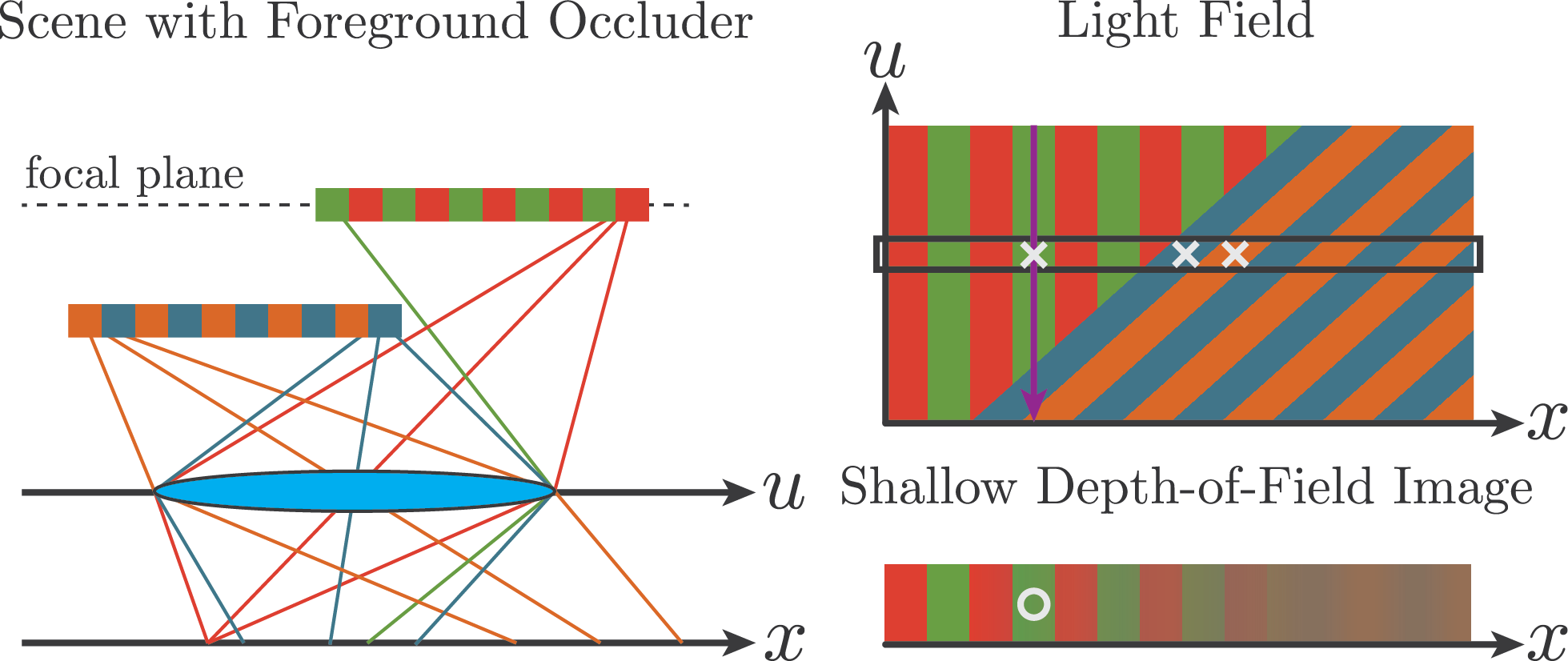} 
\caption{Our compositional aperture rendering function may not correctly render foreground occluders. On the left, we visualize an example scene layout where the green-red plane is in focus, and is occluded by the orange-blue plane. In the light field of this scene, each point on the green-red plane lies along a vertical line and each point on the orange-blue plane lies along a line with a positive slope. A single pixel in the rendered shallow depth-of-field image (white circle on the bottom right) is computed by integrating the light field along the $u$ dimension (vertical purple arrow). That pixel is the sum of green, orange, and blue non-adjacent pixels (white x's) in the input all-in-focus image (denoted by the black box), and this can be difficult to model by blending disk-blurred versions of the input all-in-focus image.
}
\label{fig:occlusions}
\end{center}
\vspace{-0.2in}
\end{figure}

While the light field aperture rendering function correctly models the light field within a camera to render a shallow depth-of-field image, it suffers from the drawback that its computational cost scales quadratically with the width of the defocus blur that it can render. 

To alleviate this issue, we propose another differentiable aperture rendering function whose computational complexity scales linearly with the width of the defocus blur that it can render. 
Instead of simulating the camera's light field to render the shallow depth-of-field image, this function models the rendering process as a depth-dependent blended composition of copies of the input all-in-focus image, each blurred with a differently sized disk-shaped kernel.

This compositional rendering function takes as input an all-in-focus image and a probabilistic depth map similar to those used in~\cite{Flynn16, Xie16}.
This probabilistic depth map $P(\mathbf{x},d;I)$ can be thought of as a per-pixel probability mass function defined over discrete disparities $d$.
We associate each of these discrete disparities with a disk blur kernel corresponding to the defocus blur for a scene point at that disparity.
The disparity associated with a blur kernel that is a delta function represents the focal plane, and the blur kernel diameter increases linearly with the absolute difference in disparity from that plane. 
We render the shallow depth-of-field image $\hat{S}_c(\mathbf{x}; I, {\hat d})$ focused at depth $\hat d$  by first shifting the probabilities so that the plane of $d = \hat d$ is associated with a delta function blur kernel, blurring the input all-in-focus image $I$ with each of the disk kernels, and then taking a weighted average of these blurred images using the values in the probabilistic depth map as weights:
% \begin{equation}
% \label{eq:conv_render}
% \hat{S}(\mathbf{x};d)=\sum_{z}P(\mathbf{x},z-d)\parens{I\parens{\mathbf{x}}*k\parens{\mathbf{x},z}}
% \end{equation}
\begin{equation}
\label{eq:conv_render}
\hat{S}_c(\mathbf{x}; I, {\hat d}) = \sum_{d}P(\mathbf{x},d - {\hat d}; I)\parens{I\parens{\mathbf{x}}*k\parens{\mathbf{x}, d}}
\end{equation}
where $*$ is convolution and $k(\mathbf{x}, d)$ is the disk blur kernel associated with depth plane $d$:
% \begin{equation}
%     k\parens{ \parens{x, y}, d} = \left[ (x^2 + y^2) \leq d^2 \right]
% \end{equation}
% or:
\begin{equation}
    k\parens{ \mathbf{x}, d} = \left[ \norm{\mathbf{x}}_2^2 \leq d^2 \right]
\end{equation}
where Iverson brackets represent an indicator function.

% If we consider the depth at each pixel as an independent random variable whose domain is the set of discrete depths, then we can also consider the value of the shallow depth-of-field image at each pixel as an independent random variable whose domain is the set of pixel values in the disk-blurred versions of the input image. 
% Our ``compositional'' aperture rendering function can be thought of as simply taking the expectation of the shallow depth-of-field image:
% \begin{equation}
% \label{eq:expectation}
% \hat{S}(\mathbf{x};d)
% =
% \mathbb{E}_{P(\mathbf{x},z-d)}\left[I(\mathbf{x})*k(\mathbf{x},z)\right]
% \end{equation}

Our compositional aperture rendering function only needs to store as many intermediate images as there are discrete depth planes, so its computational cost scales linearly with the diameter of the width of the defocus blur it can render. 
However, this increase in efficiency comes with a loss in modelling capability. More specifically, this compositional model may not correctly render the appearance of occluders closer than the focus distance. Figure~\ref{fig:occlusions} illustrates that the correct shallow depth-of-field image in a scene with a foreground occluder contains pixels that are actually the sum of non-adjacent pixels in the input all-in-focus image, so the compositional model, which is restricted to blending disk-blurred versions of the input image, may not be able to synthesize this effect in all scenes.

\section{Monocular Depth Estimation}
\label{sec:depth_estimation}

We integrate our differentiable aperture rendering functions into CNN pipelines to train functions for monocular depth estimation using aperture effects as supervision. 
The input to the full network is a single RGB all-in-focus image, and we train a CNN to predict the scene depths that minimize the difference between the ground-truth shallow depth-of-field images and those rendered by our differentiable aperture rendering functions. 
Figure~\ref{fig:pipelines} visualizes the full machine learning pipeline for each of our rendering functions.
Please refer to our supplementary materials for detailed descriptions of the CNN architectures.

\subsection{Using Light Field Aperture Rendering}

To incorporate our light field aperture rendering function into a pipeline for learning monocular depth estimation, we use a CNN $f(\cdot)$ with parameters $\boldsymbol\theta_\ell$ and the bilateral solver~\cite{Barron16} to predict a depth map $Z(\mathbf{x};I)$ from the input all-in-focus image $I(\mathbf{x})$:
\begin{equation}
\label{eq:depth_cnn}
Z(\mathbf{x};I)=\mathrm{BilateralSolver}(f_{\boldsymbol\theta_\ell}(I(\mathbf{x}))).
\end{equation} 

This results in a depth map that is smooth within similarly-colored regions and whose edges are tightly aligned with edges in the input all-in-focus image.
We use the input all-in-focus image as the bilateral space guide, and its spatial gradient magnitudes as the smoothing confidences. 
The output of the bilateral solver is differentiable with respect to the input depth map and the backward pass is fast, so we are able to integrate it into our learning pipeline and backpropagate through the solver when training. Finally, we pass this smoothed depth map and the input all-in-focus image to our light field aperture rendering functions to render a shallow depth-of-field image. 

We would like to treat $Z(\mathbf{x};I)$ as the output depth map of our monocular depth estimation system. 
Therefore, we restrict the depth expansion network $g_{\boldsymbol\theta_e}(\cdot)$ to the tasks of warping this depth map to other views and predicting the depths of occluded pixels. 
We accomplish this by regularizing the views in the depth maps predicted by $g_{\boldsymbol\theta_e}(\cdot)$ to be close to warped versions of $Z(\mathbf{x};I)$:
\begin{equation}
\label{eq:depth_reg}
\mathcal{L}_{d} \parens{D \parens{ \mathbf{x},\mathbf{u} } }
=
\norm{ D \left( \mathbf{x},\mathbf{u} \right) - Z \left( \mathbf{x} + \mathbf{u} Z \left( \mathbf{x};I \right) ;I \right) }_1
\end{equation}
where $\mathcal{L}_{d}$ is the ray depth regularization loss.

The parameters $\boldsymbol\theta_\ell$ and $\boldsymbol\theta_e$ for the CNNs that predict the depth map and expand it to a depth map for each view are learned end-to-end by minimizing the sum of the errors for rendering the shallow depth-of-field image and the ray depth regularization loss for all training tuples:
\begin{equation}
\resizebox{1.0\linewidth}{!}{$
\displaystyle
\min_{ \{ \hat{d}_i \}, \boldsymbol\theta_\ell,\boldsymbol\theta_e} \sum_i \left( \norm{\hat{S}_\ell \left( \mathbf{x}; I_i, \hat d_i \right) - S_i \left( \mathbf{x} \right) }_1+\lambda_d\mathcal{L}_d\parens{D_i \parens{ \mathbf{x},\mathbf{u} }} \right)
\label{eq:depth_loss_phys}
$}
\end{equation}
where ${I_i,S_i}$ is the i-th training tuple, consisting of an all-in-focus image $I(\mathbf{x})$ and a ground truth shallow depth-of-field image $S(\mathbf{x})$, and $\lambda_d$ is the ray depth regularization loss weight. We also minimize over the focal plane distances $\hat{d}_i$ for each training example, so our algorithm does not require the in-focus disparity to be given.
This also sidesteps the difficult problem of recording $\hat d_i$ for each image during dataset collection, which would require control over image metadata and knowledge of the camera and lens parameters.

%\rahulcom{It's hard to understand why you are minimizing over $\hat{d}_i$ without reading Learning the Focal Plane Distance section. How about having the following equation here:
%\begin{equation}
%\min_{ \boldsymbol\theta_c,\boldsymbol\theta_e} \sum_i \left( \norm{\hat{S}_\ell \left( \mathbf{x}; I_i, d_i \right) - S_i \left( \mathbf{x}, d_i \right) }_1+\lambda_d\mathcal{L}_d \right)
%\end{equation}
%and then refining it to your version in the later Section?}

\subsection{Using Compositional Aperture Rendering}

To use our compositional aperture rendering function in a pipeline for learning monocular depth estimation, we have the depth estimation CNN $f_{\boldsymbol\theta_c}(\cdot)$ output values over $n$ discrete depth planes instead of just a single depth map:
\begin{equation}
P\parens{\mathbf{x}, d; I} = f_{\boldsymbol\theta_c}(I(\mathbf{x})).
\end{equation}
The predicted values for each pixel are then normalized by a $\mathit{softmax}$, so we can consider $P\parens{\mathbf{x}, d; I}$ to be a probabilistic depth map composed of a probability mass function (PMF) that sums to $1$ for each pixel. We pass $P\parens{\mathbf{x}, d; I}$ and the input image $I$ to our compositional aperture rendering function to render a shallow depth-of-field image. 

Unlike the light field aperture rendering function, this pipeline does not contain a depth expansion network, so we train the parameters of the depth prediction network by minimizing the sum of the errors for rendering the shallow depth-of-field image as well as a total variation regularization of the probabilistic depth maps, for all training tuples:
\begin{equation}
\resizebox{1.0\linewidth}{!}{$
\displaystyle
\min_{\{ \hat{d}_i \}, \boldsymbol\theta_c} \sum_i \left( \norm{\hat{S}_c(\mathbf{x};I_i, \hat{d}_i )-S_i(\mathbf{x})}_1+\sum_d\lambda_{tv}\norm{\nabla P\parens{\mathbf{x}, d; I_i}}_1 \right)
$}
\label{eq:depth_loss_compositional}
\end{equation}
where $\nabla$ indicates the partial derivatives (finite differences [-1,1] and [-1;1]) in $x$ and $y$ of each channel of $P(\cdot)$.
% \begin{equation}
% \hat{S}_c(\mathbf{x}; I, {\hat d}) = \sum_{d}P(\mathbf{x}, d - {\hat d}; I)\parens{I\parens{\mathbf{x}}*k\parens{\mathbf{x}, d}}
% \end{equation}
% So then instead of $P_i(\mathbf{x}, d)$ we'd have
% \begin{equation}
% P\parens{\mathbf{x}, d; I_i}
% \end{equation}

% \subsection{Learning the Focal Plane Distance}

% Both aperture rendering functions require the focal plane distance $\hat{d}_i$ to produce a shallow depth-of-field rendering for image $i$
% Recording this value during dataset collection is difficult, but even if we could obtain it, the value of $\hat{d}_i$ within our pipeline is not defined in terms of physical metric distance. 
% Instead, $d$ represents disparity across the camera aperture, and relating it to metric distance is difficult without full knowledge of the camera and lens parameters. 

% To address this we create a variable for the value of $d$ for each training example, and we optimize these variables along with the network parameters during training. At test time, the user can input various value of $d$ into the rendering function to generate shallow depth-of-field images focused to different distances.

\subsection{Depth Ambiguities}

Training a monocular depth estimation algorithm by direct regression from an image to a depth map is straightforward and unambiguous, but ambiguities arise when relying on indirect sources of depth information.
% When training a monocular depth estimation algorithm it is important to consider the ambiguities inherent in the form of supervision used for training.
% For example, there are no training ambiguities when using ground-truth depth as supervision, since we are simply training the network to regress to the ground-truth values.
E.g., if we use images from an alternate viewpoint as supervision~\cite{Garg16,Godard17,Xie16} there is an ambiguity for image regions whose appearance is constant or repetitive along epipolar line segments --- many predicted depths would result in a perfect match in the alternate image.
This can be remedied by training with pairs that have different relative camera positions, so that the baseline and orientation of the epipolar lines varies across the training examples~\cite{Zhou17}.

Aperture supervision suffers from two main ambiguities. First, there is a sign ambiguity for the depths that correctly render a given shallow depth-of-field image: any out-of-focus scene point, in the absence of occlusions, could be located in front of or behind the focal plane. Second, the depth is ambiguous within constant image regions, which look identical with any amount of defocus blur.
We address the first ambiguity by ensuring a diversity of focus in our datasets: objects appear at a variety of distances relative to the focal plane.
We address the second ambiguity by applying a bilateral solver to our predicted depth maps, using the gradient magnitude of the input image as the confidence.
This doesn't remove the ambiguity in the data, but it effectively encodes a prior that depth predictions at image edges are more trustworthy than those in smooth regions.

\section{Results}

We evaluate the performance of aperture supervision with our two differentiable aperture rendering functions for training monocular depth estimation models.
Evaluating performance on this task is challenging, as we are not aware of any prior work that addresses this task.
We therefore compare our results to state-of-the-art methods that use different forms of supervision.
Since ground truth depth is not available in our training datasets, we qualitatively compare the predicted scene depths in Figures~\ref{fig:lytro_qual} and \ref{fig:slr_qual}, and quantitatively compare the shallow depth-of-field images rendered with our algorithm to those rendered using scene depths predicted by the baseline techniques in Tables~\ref{table:lytro} and \ref{table:slr}.
We visualize the probabilistic depths from our compositional rendering model by taking the pixel-wise mode of each PMF and smoothing this projection with the bilateral solver.

\subsection{Baseline Methods}

We use Laina \etal~\cite{Laina16} as a representative state-of-the-art technique for training a network to predict scene depths using ground truth depths as supervision.
We use their model trained on the NYU Depth v2 dataset~\cite{Silberman12}, which consists of aligned pairs of RGB and depth images taken with the Microsoft Kinect V1.
This model predicts metric depths as opposed to disparities, so naively treating the output of this model as disparity would be unfair to this work. 
To be maximally generous to this baseline, we fit a piecewise linear spline to transform their predicted depths to minimize the squared error with respect to our light field model's disparities.
The ``warped individually'' baseline was computed by fitting a 5-knot linear spline for each image being evaluated. 
The ``warped together'' baseline was computed by fitting a single 17-knot linear spline to the set of all pairs of depth maps.

Our ``Multi-View Supervision'' baseline is intended to evaluate the differences between using aperture effects and view synthesis as supervision.
We train a monocular depth prediction network that is identical to that used in our light field rendering pipeline, including the bilateral solver. 
As is typical in multi-view supervision, our loss function is the $L_1$ error between the input image and an image from an alternate viewpoint warped into the viewpoint of the input image according to the predicted depth map. 
To perform a fair comparison where every model component is held constant besides the type of supervision, we use an image taken from a viewpoint at the edge of the light field camera's aperture as the alternate view, so the disparity between the two images used for multi-view supervision is equal to the radius of the defocus blur used for our aperture supervision algorithms.
We consider these results as representative of state-of-the-art monocular depth estimation algorithms that use multi-view supervision for training~\cite{Garg16,Godard17,Xie16,Zhou17}.

Our ``Image Regression'' baseline is a network that is trained to directly regress to a shallow depth-of-field image, given the input all-in-focus image and the desired aperture size and focus distance. We append the aperture size and focus distance to the input image as additional channels, and use the same architecture as our depth estimation network.

\subsection{Light Field Dataset Experiments}

\begin{figure*}
\begin{center}
\newcommand{\width}{1.0\linewidth}
\includegraphics[width=\width]{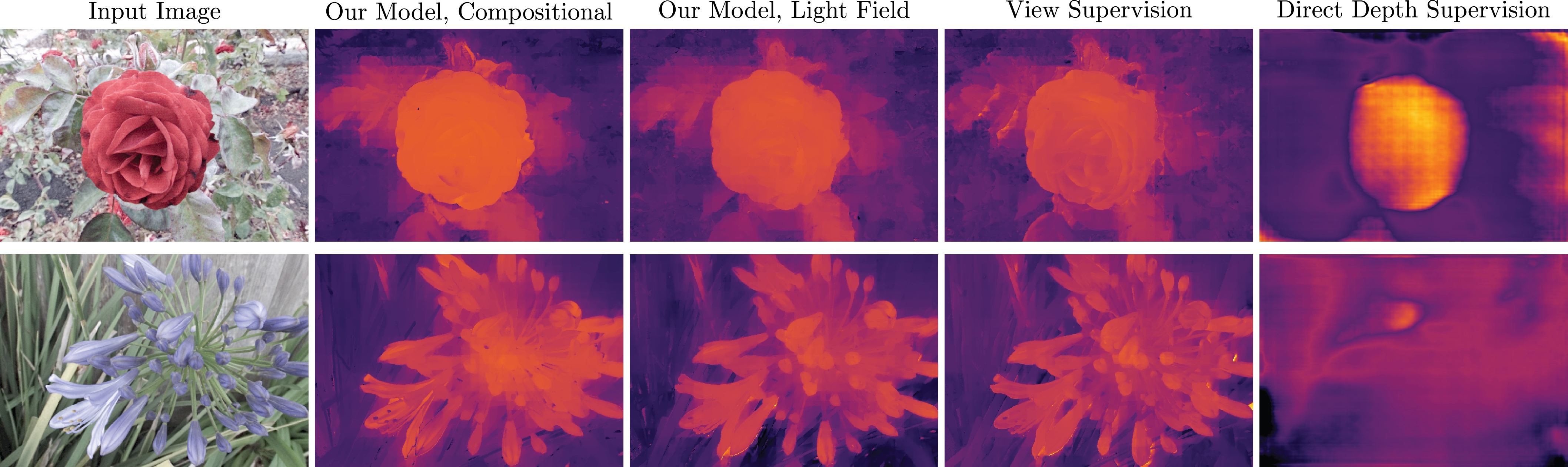} 
\caption{A qualitative comparison of monocular depth estimation results on images from the test set of our light field experiments. 
Our aperture supervision models are able to estimate high-quality detailed depths. 
The depths estimated by a network trained with multi-view supervision are reasonable, but typically have artifacts around occlusion edges.}
\label{fig:lytro_qual}
\end{center}
\vspace{-0.15in}
\end{figure*}

\begin{figure*}
\begin{center}
\newcommand{\width}{1.0\linewidth}
\includegraphics[width=\width]{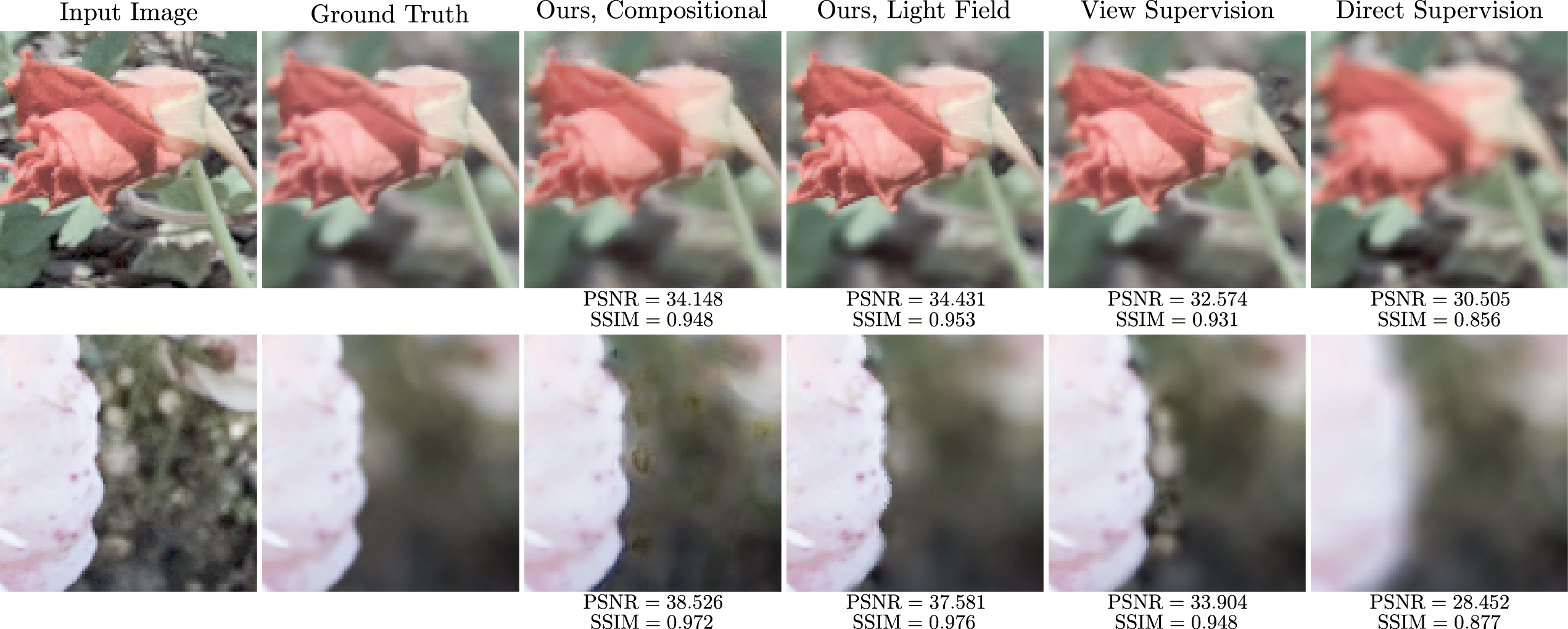} 
\caption{A quantitative and qualitative comparison of crops from rendered shallow depth-of-field images from the test set of our light field experiments. 
The images rendered using depths predicted by our models trained with aperture supervision closely match the ground truth. 
Images rendered using depths trained by multi-view supervision contain false edges and artifacts near occlusion edges, and images rendered using depths trained by direct depth supervision do not contain any reasonable depth-of-field effects.}
\label{fig:lytro_qual_bokeh}
\end{center}
\vspace{-0.2in}
\end{figure*}

We use a recently-introduced dataset~\cite{Srinivasan17} of light fields of flowers and plants, taken with the Lytro Illum camera using a focal length of 30 mm, to evaluate our aperture supervision methods and compare them to the baselines of image regression, direct depth supervision, and multi-view supervision.
The all-in-focus and shallow depth-of-fields that we synthesize from these light fields are equivalent to images taken with aperture sizes $\sfrac{f}{28}$ and $\sfrac{f}{2.3}$.
We randomly partition this dataset into a training set of 3143 light fields, and a test set of 300 light fields.
Table~\ref{table:lytro} shows that our model quantitatively outperforms all baseline techniques. Figure~\ref{fig:lytro_qual} visualizes example monocular depth estimation results. Aperture supervision with our two differentiable rendering functions produces high-quality depths, while depth maps estimated by multi-view supervision networks contain artifacts at occlusion edges. As demonstrated in Figure~\ref{fig:lytro_qual_bokeh}, these artifacts in the depth maps cause false edges and distracting textures in the rendered shallow depth-of-field images, while our rendered images contain natural and convincing synthetic defocus blur.

%\barroncom{One nice thing about this dataset is that it's outdoors during the day, which is an environment where most direct depth sensors (structured light or time-of-flight) would fail due to the significant ambient light. Might be a cool thing to point out that even if we had a mobile Kinect or anything short of a LIDAR, this wouldn't work. This could be framed in a way that hedges against criticism of using the im2depth model--- Reviewer sez: "Why didn't you compare against Kinect images of flowers?" and we say "because you can't collect direct depth data of them"}

\subsection{DSLR Dataset Experiments}

\begin{figure}
\begin{center}
\includegraphics[width=1.0\linewidth]{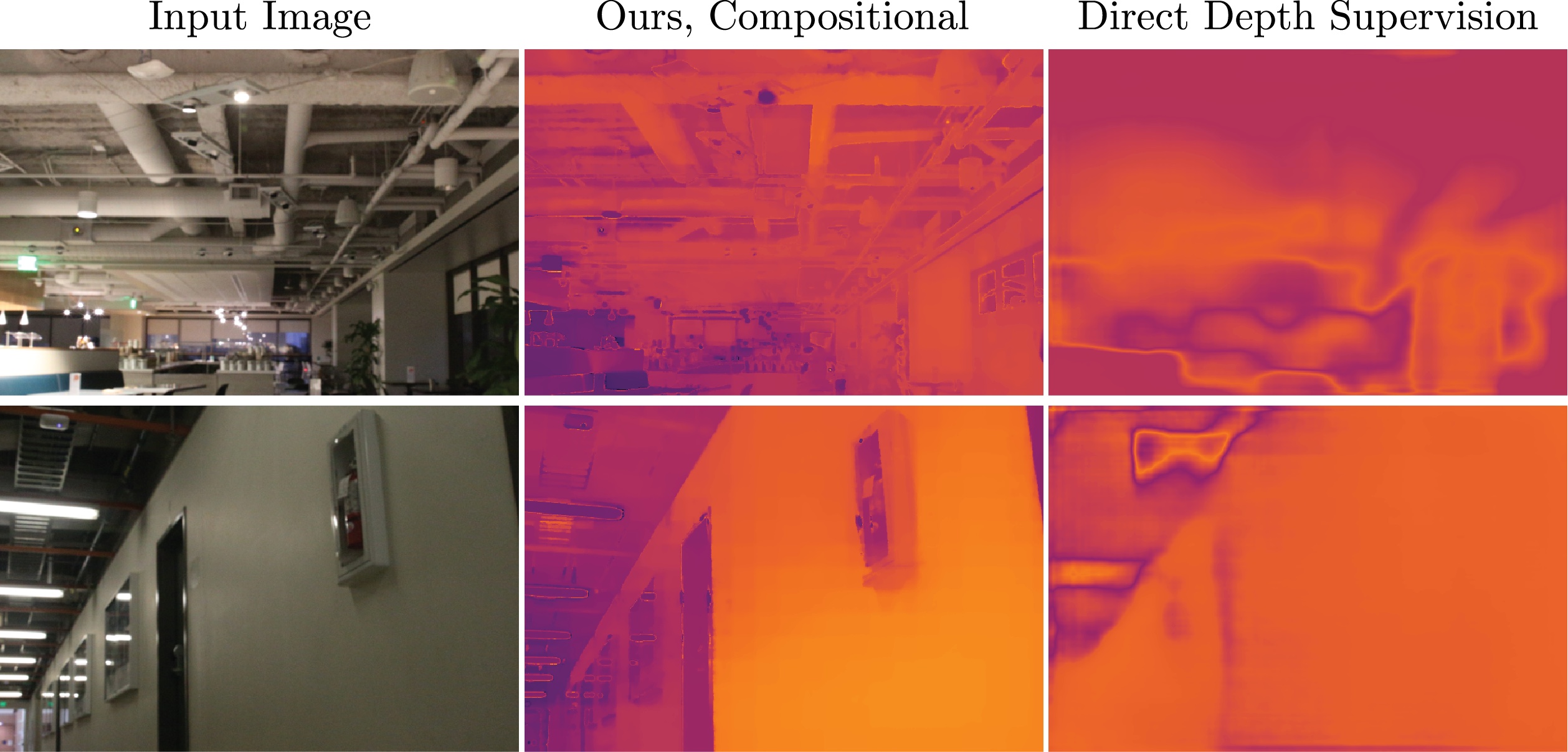} 
\caption{A qualitative comparison of monocular depth estimation results from the test set of our DSLR dataset experiments.
Our aperture supervision model is able to estimate more detailed depth maps than the direct depth supervision baseline.}
\label{fig:slr_qual}
\end{center}
\vspace{-0.2in}
\end{figure}

To further validate aperture supervision, we gathered a dataset with a Canon 5D Mark III camera, consisting of images of 758 scenes taken with a focal length of $24$mm. 
For each scene, we captured images from the same viewpoint, focused at $0.5$m and $1$m, each taken with $\sfrac{f}{14}$ and $\sfrac{f}{3.5}$ apertures.
This dataset was collected such that it contains the same sorts of indoor scenes as the NYU Depth v2 dataset~\cite{Silberman12}, in an effort to be as generous as possible towards our direct depth supervision baseline.
We randomly partition this dataset into a training set of 708 tuples, each containing a single $\sfrac{f}{14}$ image and the corresponding two $\sfrac{f}{3.5}$ images, and a test set of 50 tuples.
Since this dataset does not contain images taken from alternate viewpoints, we only compare the depth estimation results of our methods to those using direct depth supervision.
Table~\ref{table:slr} shows that our model quantitatively outperforms the direct depth supervision and image regression baselines, and Figure~\ref{fig:slr_qual} demonstrates that our trained algorithm is able to estimate much sharper and higher-quality depths than direct depth supervision.
The dearth of applicable baseline techniques for this task highlights the value of our technique. 
There are no techniques that we are aware of which can take advantage of our training data, and there are few ways to otherwise train a monocular depth-estimation algorithm.

\subsection{Training Details}

We synthesize light fields with $12\times12$ views in our light field rendering function for the light field dataset experiments, and $4\times4$ views for the DSLR dataset experiments.
When using our compositional aperture rendering function, we use $n=31$ depth planes, with $d \in [-15,15]$.
Our regularization hyperparameters are $\lambda_d=0.1$ and $\lambda_{tv}=10^{-10}$.
We use the Adam optimizer~\cite{KingmaB14} with a learning rate of $10^{-4}$ and a batch size of $1$, and train for 240K iterations.  
All of our models were implemented in Tensorflow~\cite{tensorflow2015}.

\begin{table}[!]
\begin{center}
\resizebox{3.25in}{!}{
\Huge
\begin{tabular}{ l | cccc}
Algorithm & PSNR $d_1$ & SSIM $d_1$ & PSNR $d_2$ & SSIM $d_2$ \\
\hline
                   Image Regression & 24.60 $\pm$ 1.39  & 0.895 $\pm$ 0.045  & 24.49 $\pm$ 1.31  & 0.888 $\pm$ 0.047 \\
 \cite{Laina16} Warped Individually & 31.95 $\pm$ 2.17  & 0.909 $\pm$ 0.034  & 31.50 $\pm$ 2.19  & 0.903 $\pm$ 0.040 \\
     \cite{Laina16} Warped Together & 31.59 $\pm$ 2.79  & 0.895 $\pm$ 0.051  & 31.39 $\pm$ 2.28  & 0.904 $\pm$ 0.041 \\
                     Multi-View Supervision & 34.49 $\pm$ 1.87  & 0.960 $\pm$ 0.017  & 34.36 $\pm$ 1.72  & 0.956 $\pm$ 0.017 \\
             Our Model, Light Field & 36.68 $\pm$ 2.03  & \cellcolor{Yellow} 0.967 $\pm$ 0.016  & 35.58 $\pm$ 1.86  & 0.961 $\pm$ 0.015 \\
           Our Model, Compositional & \cellcolor{Yellow} 36.90 $\pm$ 2.11  & 0.966 $\pm$ 0.016  & \cellcolor{Yellow} 35.76 $\pm$ 1.97  & \cellcolor{Yellow} 0.963 $\pm$ 0.016 \\

\end{tabular}
}
\vspace{1mm}
\caption{A quantitative comparison on the 300-image test set from our light field experiments. We report the mean and standard deviation PSNR and SSIM of synthesized $\sfrac{f}{2.3}$ images for two target focus distances, $d_1$ (focused on the subject flower) and $d_2$ (focused to the light field's maximum refocusable depth).
\label{table:lytro}
}
\vspace{0.1in}
\resizebox{3.25in}{!}{
\Huge
\begin{tabular}{ l | cccc}
Algorithm & PSNR $d_1$ & SSIM $d_1$ & PSNR $d_2$ & SSIM $d_2$ \\
\hline
                   Image Regression & 22.26 $\pm$ 4.89  & 0.958 $\pm$ 0.022  & 20.93 $\pm$ 3.09  & 0.851 $\pm$ 0.046 \\
 \cite{Laina16} Warped Individually & 28.31 $\pm$ 4.36  & 0.928 $\pm$ 0.040  & 32.52 $\pm$ 3.34  & 0.953 $\pm$ 0.030 \\
     \cite{Laina16} Warped Together & 28.54 $\pm$ 4.42  & 0.933 $\pm$ 0.034  & 32.52 $\pm$ 3.33  & 0.953 $\pm$ 0.030 \\
             Our Model, Light Field & \cellcolor{Yellow} 35.39 $\pm$ 3.80  & 0.976 $\pm$ 0.011  & 33.01 $\pm$ 3.59  & 0.955 $\pm$ 0.028 \\
           Our Model, Compositional & 33.87 $\pm$ 5.09  & \cellcolor{Yellow} 0.983 $\pm$ 0.010  & \cellcolor{Yellow} 33.28 $\pm$ 3.25  & \cellcolor{Yellow} 0.962 $\pm$ 0.025 \\

\end{tabular}
}
\vspace{1mm}
\caption{
A quantitative comparison on the 50-image test set from our DSLR experiments. We report the mean and standard deviation PSNR and SSIM of synthesized $\sfrac{f}{3.5}$ images for two target focus distances, $d_1=0.5\text{m}$ and $d_2=1\text{m}$.
\label{table:slr}
}
\end{center}
\vspace{-0.22in}
\end{table}

\section{Conclusions}

We have presented a new way to train machine learning algorithms to predict scene depths from a single image, using camera aperture effects as supervision.
By including a differentiable aperture rendering function within our network, we can train a network to regress from a single all-in-focus image to the depth map that best explains a paired shallow depth-of-field image.
This approach produces more accurate synthetic defocus renderings than other approaches due to the supervisory signal being consistent with the desired task, and also relies on training data from a single conventional camera that is easier to collect than depth-sensor- or stereo-based approaches.
% By training networks that contain differentiable aperture rendering functions to estimate the depth maps that best explain shallow depth-of-field images, we produce an model that produces more accurate synthetic defocus rendering than other approaches, while relying on training data that is easier to acquire.
% for the computational photography task of synthetic defocus, and we simplify the task of dataset collection by only requiring a single camera.
% This allows us to train networks to estimate the depth maps that best render shallow depth-of-field images taken with larger camera apertures, so our depth estimation is trained end-to-end for the computational photography task of synthetic defocus.
% To enable machine learning pipelines to train with aperture supervision, we introduced two differentiable aperture rendering functions.
Our model has two variants, each with its own differentiable aperture rendering function.
Our ``light field'' model uses a continuous-valued depth map and an explicit simulation of light rays within a camera to produce more geometrically-accurate results, but with a computational cost that scales quadratically with respect to the maximum synthetic blur size.
Our ``compositional'' model uses a discrete per-pixel PMF over depths and a filter-based rendering approach to achieve a linear complexity with respect to blur size, but uses a probabilistic depth estimate that may not be trivial to adapt to different tasks.
Aperture supervision represents a novel and effective form of supervision that is complementary to and compatible with existing forms of supervision (such as multi-view supervision or direct depth supervision) and may enable the explicit geometric modelling of image formation in other machine learning pipelines.

% The first is a light field model that simulates the light rays in a camera in a geometrically correct manner and enables the network to predict a continuous-valued depth map, but has a computational cost that scales quadratically with the maximum blur size.
% The second is a compositional model that requires the depth estimation network to predict a per-pixel PMF over a discrete set of depths, which may be more difficult to use than a continuous-valued depth map for other tasks. However, the compositional model has a computatational cost that scales just linearly with the maximum blur size.

% Furthermore, aperture supervision is complementary to multi-view supervision, and our differentiable aperture rendering functions could be used along with depth-based warping to train a monocular depth estimation network with a dataset containing images of a scene taken from different viewpoints as well as different aperture settings.

% We believe that this work opens up exciting opportunities for differentiably modelling image formation as well as combining multiple types of supervision into a common framework for training machine learning algorithms to estimate geometry from images.

\clearpage

{\small
\bibliographystyle{ieee}
\bibliography{mainbib}
}

\clearpage

\section{Supplementary Materials}

\subsection{Network Architectures}

Here, we provide detailed descriptions of the convolutional neural network (CNN) architectures used in our method.

$C_{i}S_{j}D_{k}$ denotes a convolution layer with $i$ 3x3 filters, a spatial stride of $j$ in each dimension, and a dilation rate of $k$.
Additionally, $E$ denotes an exponential linear unit activation function~\cite{Clevert16} and $I$ denotes instance normalization~\cite{Ulyanov16}.
Finally, $R$ denotes a residual connection where the current tensor is added to the tensor output from the previous instance normalization layer.

For our light field dataset experiments, the monocular depth estimation CNN $f_{\boldsymbol\theta_d}(\cdot)$ contains 12 convolutional layers structured as:
% $C_{8}S_{1}D_{1}-E-I-C_{32}S_{1}D_{1}-E-I-C_{64}S_{1}D_{1}-E-I-C_{128}S_{1}D_{1}-E-I-C_{128}S_{1}D_{2}-E-I-R-C_{128}S_{1}D_{4}-E-I-R-C_{128}S_{1}D_{8}-E-I-R-C_{128}S_{1}D_{16}-E-I-R-C_{128}S_{1}D_{32}-E-I-R-C_{64}S_{1}D_{1}-E-I-C_{32}S_{1}D_{1}-E-I-C_{1}S_{1}D_{1}$.
\mathchardef\mhyphen="2D % Define a "math hyphen"
\begin{align}
& C_{8}S_{1}D_{1} \mhyphen E \mhyphen I \mhyphen C_{32}S_{1}D_{1} \mhyphen E \mhyphen I \mhyphen C_{64}S_{1}D_{1} \mhyphen E \mhyphen I \mhyphen \nonumber \\
& C_{128}S_{1}D_{1} \mhyphen E \mhyphen I \mhyphen C_{128}S_{1}D_{2} \mhyphen E \mhyphen I \mhyphen R \mhyphen C_{128}S_{1}D_{4} \mhyphen E \mhyphen I \mhyphen R \mhyphen  \nonumber \\
& C_{128}S_{1}D_{8} \mhyphen E \mhyphen I \mhyphen R \mhyphen C_{128}S_{1}D_{16} \mhyphen E \mhyphen I \mhyphen R \mhyphen C_{128}S_{1}D_{32} \mhyphen E \mhyphen I \mhyphen R \mhyphen \nonumber \\
& C_{64}S_{1}D_{1} \mhyphen E \mhyphen I \mhyphen C_{32}S_{1}D_{1} \mhyphen E \mhyphen I \mhyphen C_{1}S_{1}D_{1}.
\end{align}
For our SLR dataset experiments, the monocular depth estimation CNN $f_{\boldsymbol\theta_d}(\cdot)$ contains 12 convolutional layers structured as:
% $C_{8}S_{1}D_{1}-E-I-C_{32}S_{1}D_{1}-E-I-C_{64}S_{1}D_{1}-E-I-C_{128}S_{1}D_{1}-E-I-C_{128}S_{1}D_{2}-E-I-R-C_{128}S_{1}D_{4}-E-I-R-C_{128}S_{1}D_{8}-E-I-R-C_{128}S_{1}D_{16}-E-I-R-C_{128}S_{1}D_{32}-E-I-R-C_{64}S_{1}D_{1}-E-I-C_{32}S_{1}D_{1}-E-I-C_{1}S_{1}D_{1}$.
\begin{align}
& C_{4}S_{2}D_{1} \mhyphen E \mhyphen I \mhyphen C_{8}S_{2}D_{1} \mhyphen E \mhyphen I \mhyphen C_{16}S_{1}D_{1} \mhyphen E \mhyphen I \mhyphen  \nonumber \\
& C_{64}S_{1}D_{1} \mhyphen E \mhyphen I \mhyphen C_{64}S_{1}D_{2} \mhyphen E \mhyphen I \mhyphen R \mhyphen C_{64}S_{1}D_{4} \mhyphen E \mhyphen I \mhyphen R \mhyphen  \nonumber \\
& C_{64}S_{1}D_{8} \mhyphen E \mhyphen I \mhyphen R \mhyphen C_{64}S_{1}D_{16} \mhyphen E \mhyphen I \mhyphen R \mhyphen C_{64}S_{1}D_{32} \mhyphen E \mhyphen I \mhyphen R \mhyphen  \nonumber \\
& C_{64}S_{1}D_{1} \mhyphen E \mhyphen I \mhyphen R \mhyphen C_{32}S_{1}D_{1} \mhyphen E \mhyphen I \mhyphen C_{1}S_{1}D_{1}.
\end{align}

When using our light field aperture rendering function, the output of the monocular depth estimation network is passed through a scaled $\mathit{tanh}(\cdot)$ activation function to restrict the disparities to $[-10,10]$ pixels between adjacent views. When using our compositional aperture rendering function, the number of filters in the last convolutional layer is modified to be the number of discrete depth planes $n$.

For all experiments using the light field aperture rendering function, the depth expansion CNN $g_{\boldsymbol\theta_e}(\cdot)$ contains 3 convolutional layers structured as:
\begin{align}
& C_{m}S_{1}D_{1} \mhyphen E \mhyphen I \mhyphen C_{m}S_{1}D_{1} \mhyphen E \mhyphen I \mhyphen C_{n}S_{1}D_{1}
\end{align}
where $m$ is the total number of views in the light field.

\subsection{Additional Results}

Below, we display additional qualitative results for both our light field and DSLR experiments.

\begin{figure*}
\begin{center}
\newcommand{\width}{1.0\linewidth}
\includegraphics[width=\width]{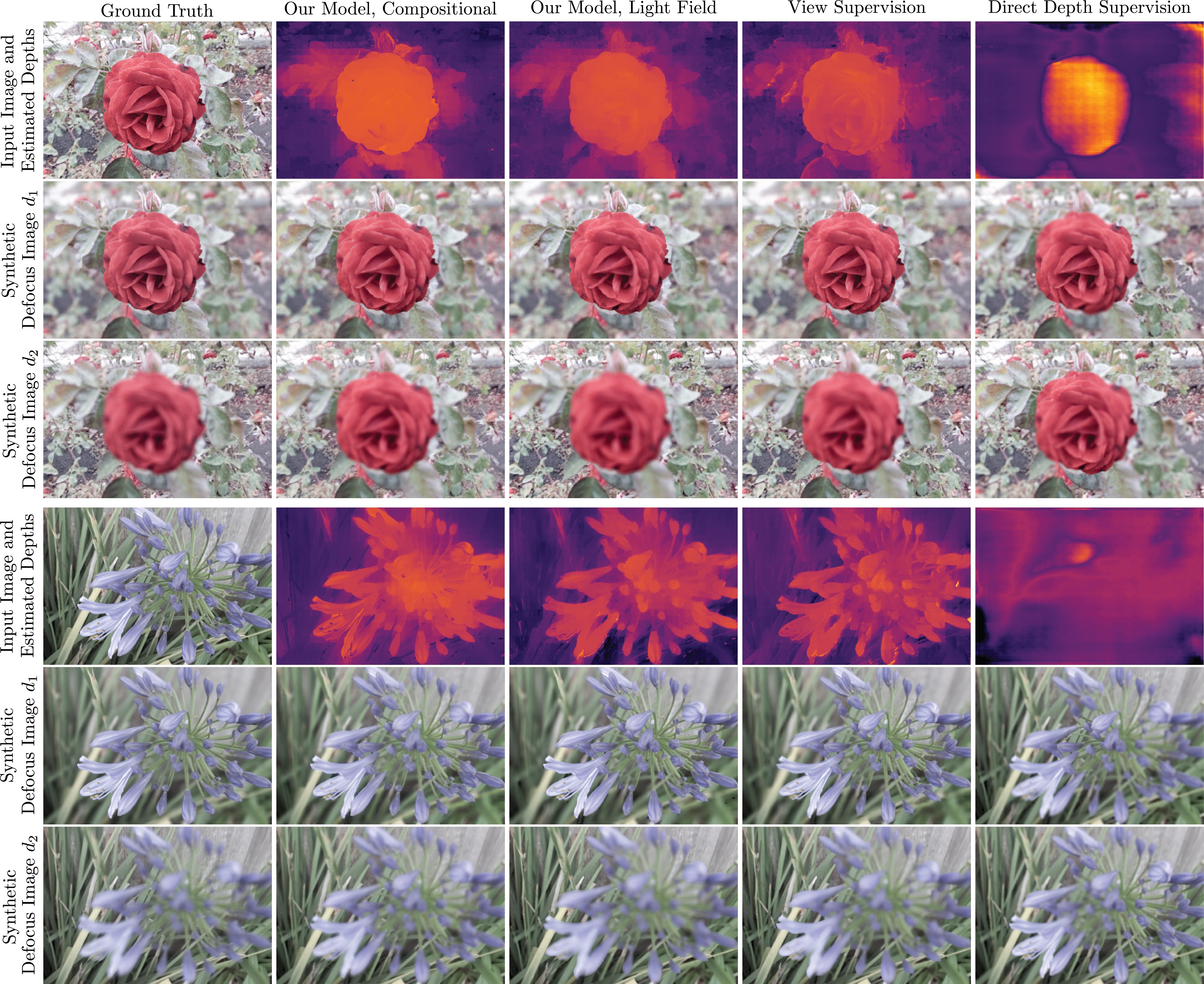} 
\caption{Additional qualitative comparison of monocular depth estimation and synthetic defocus results on images from the test set of our light field experiments. 
Our aperture supervision models are able to estimate high-quality detailed depths and render convincing shallow-depth-of-field images. 
The depths estimated by a network trained by view synthesis supervision are reasonable, but typically have artifacts around occlusion edges, causing false edges and artifacts in their rendered shallow depth-of-field images. We recommend that readers view these figures digitally and zoom in to see fine details and differences between the various methods.}
\label{fig:lytro_qual_1}
\end{center}
\end{figure*}

\begin{figure*}
\begin{center}
\newcommand{\width}{1.0\linewidth}
\includegraphics[width=\width]{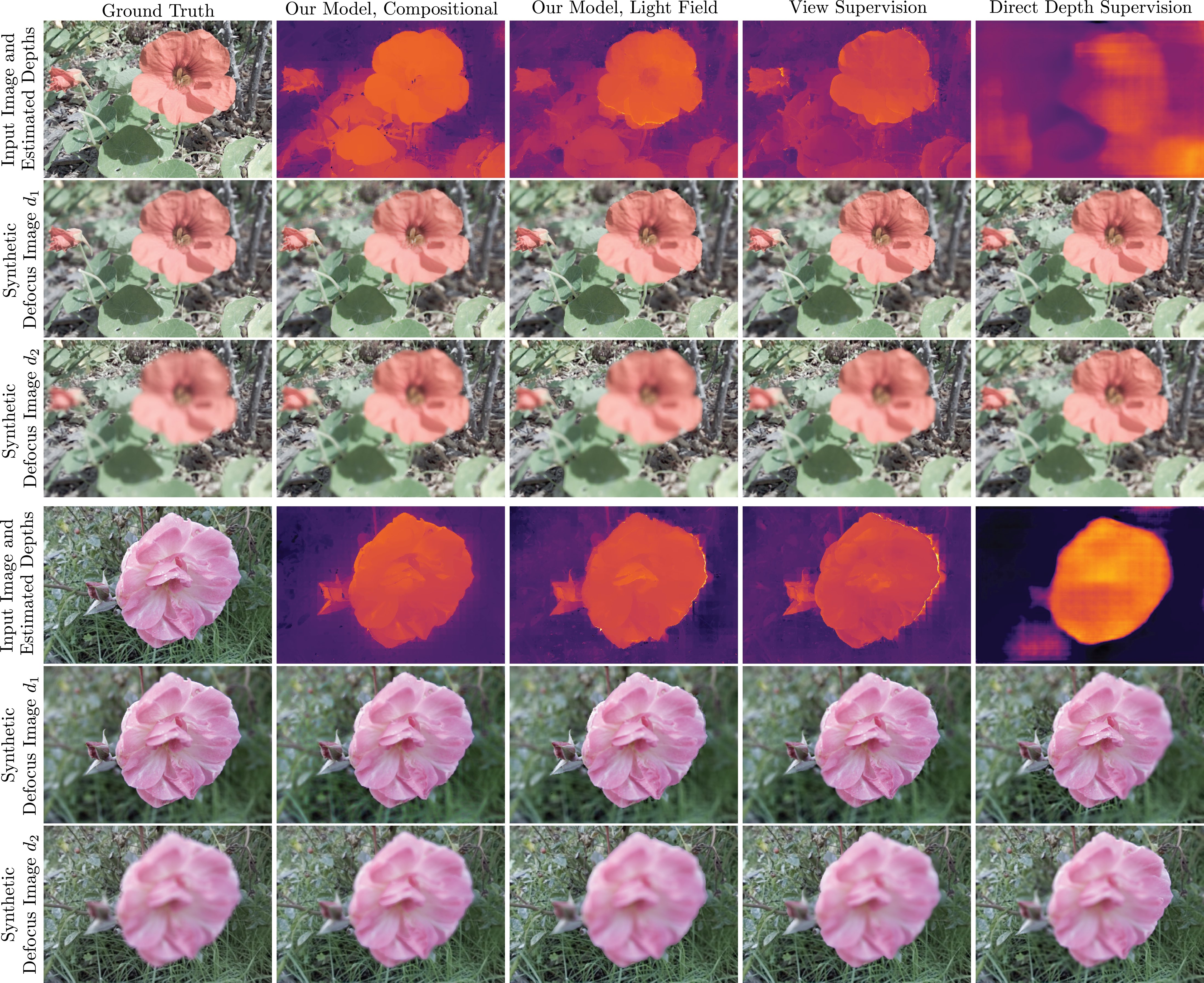} 
\caption{Additional qualitative comparison of monocular depth estimation and synthetic defocus results on images from the test set of our light field experiments. 
Our aperture supervision models are able to estimate high-quality detailed depths and render convincing shallow-depth-of-field images. 
The depths estimated by a network trained by view synthesis supervision are reasonable, but typically have artifacts around occlusion edges, causing false edges and artifacts in their rendered shallow depth-of-field images. We recommend that readers view these figures digitally and zoom in to see fine details and differences between the various methods.}
\label{fig:lytro_qual_2}
\end{center}
\end{figure*}

\begin{figure*}
\begin{center}
\newcommand{\width}{1.0\linewidth}
\includegraphics[width=\width]{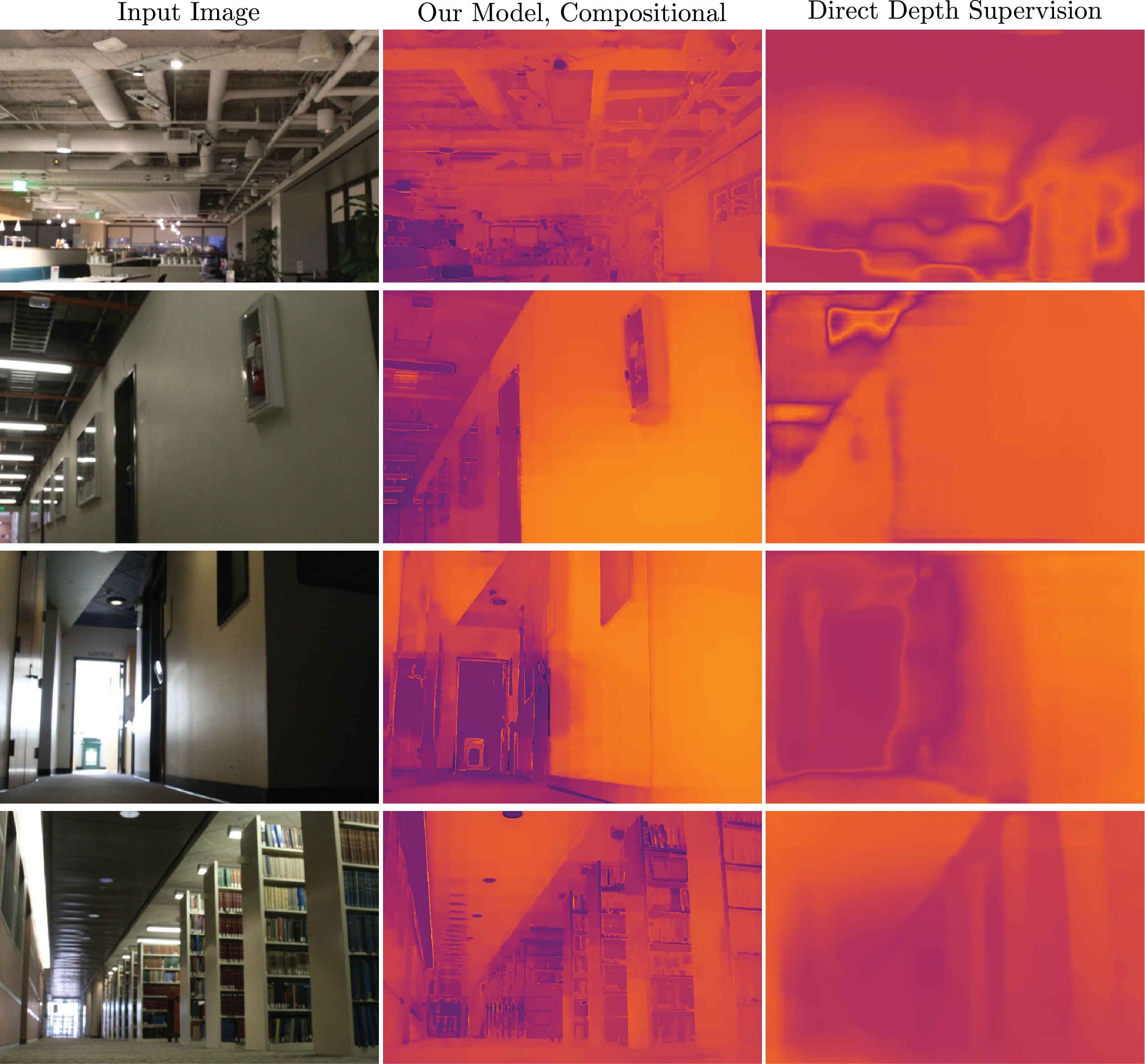} 
\caption{Additional qualitative comparison of monocular depth estimation results on images from the test set of our DSLR dataset experiments.
Our aperture supervision model is able to estimate more detailed depth maps than the direct depth supervision baseline.}
\label{fig:slr_qual_supp}
\end{center}
\end{figure*}

\end{document}